\documentclass[runningheads]{llncs}
\usepackage[T1]{fontenc}
\usepackage{graphicx}
\usepackage{colortbl}
\usepackage{pseudocode} 
\usepackage{multirow}
\usepackage{enumitem}
\usepackage{url}
\usepackage{tikz}
\usepackage{balance}
\usepackage[skip=1pt]{caption}
\usepackage{booktabs}
\usepackage{cite}

\usepackage{algorithm, algpseudocode}
\algnewcommand\algorithmicforeach{\textbf{for each}}
\algdef{S}[FOR]{ForEach}[1]{\algorithmicforeach\ #1\ \algorithmicdo}

\usepackage{pbox}

\newcommand{\mybox}[1]
{
\vspace{1.8mm}
\noindent \hspace{-1mm} 
\setlength\fboxsep{1mm}
\fbox{\parbox{\dimexpr\linewidth-2\fboxsep-2\fboxrule}{\itshape #1}}
}
\usepackage{amsmath}

\newcommand{\para}[1]{\vspace{0.1mm}\noindent\textbf{#1}.} 
\newcommand{\orion}{kumar2015learning}
\newcommand{\santoku}{kumar_demonstration_2015}

\newcommand{\f}{schleichLearningLinearRegression2016}

\newcommand{\acdc}{khamisACDCInDatabase2018}

\newcommand{\morpheusfi}{liEnablingOptimizingNonlinear2019}

\newcommand{\hadad}{alotaibiHADADLightweightApproach2021}

\newcommand{\trinity}{justoPolyglotFrameworkFactorized2021}

\begin{document}
\title{Ilargi: a GPU Compatible Factorized ML Model Training Framework}
\author{Wenbo Sun\inst{1} \and
Rihan Hai\inst{1}}
\institute{Delft University of Technology \\
\email{w.sun-2@tudelft.nl} \\
\email{r.hai@tudelft.nl}}
%
%
%
%
\maketitle              
\begin{abstract}
The machine learning (ML) training over disparate data sources traditionally involves materialization, which can impose substantial time and space overhead due to data movement and replication. Factorized learning, which leverages direct computation on disparate sources through linear algebra (LA) rewriting, has emerged as a viable alternative to improve computational efficiency. However, the adaptation of factorized learning to leverage the full capabilities of modern LA-friendly hardware like GPUs has been limited, often requiring manual intervention for algorithm compatibility. This paper introduces \emph{Ilargi}, a novel factorized learning framework that utilizes matrix-represented data integration (DI) metadata to facilitate automatic factorization across CPU and GPU environments without the need for costly relational joins. \emph{Ilargi} incorporates an ML-based cost estimator to intelligently selects between factorization and materialization based on data properties, algorithm complexity, hardware environments, and their interactions. This strategy ensures up to 8.9x speedups on GPUs and achieves over 20\% acceleration in batch ML training workloads, thereby enhancing the practicability of ML training across diverse data integration scenarios and hardware platforms. To our knowledge, this work is the very first effort in GPU-compatible factorized learning.

\end{abstract}

\vspace{-6mm}
\section{Introduction}
\label{sec:intro}
\vspace{-3mm}
The raw data for machine learning (ML) model training is often dispersed across multiple database tables or physically separate sources \cite{olteanu2020relational, kumar2017data}. These data sources typically undergo a data integration (DI) process, known as materialization, to create a unified view \cite{10.5555/1182635.1164130} for subsequent machine learning training. Though widely adopted, materialization introduces considerable overhead, both in terms of time and space, due to extensive data movement and replication during data integration.

Factorized learning  \cite{\orion, \f, \acdc, chen2017towards, MorpheusFI, DBLP:conf/icde/ChengKZ021} offers a strategy to train ML models directly over disparate sources without materialization. Factorized learning can improve model training efficiency by pushing operators in ML training  down to the data source side. 
For example, to train a linear regression model using gradient descent to predict transaction amounts (\(T\)) based on customer profiles (\(P\)) and transaction history (\(H\)), traditional approaches first materialize a table by joining \(P\) and \(H\) on customer ID. This involves substantial data movement and preprocessing time, especially for large datasets. Factorized learning, in contrast, avoids materialization by leveraging additive linear algebra rewriting to decompose the training process. It trains sub-models on \(P\) and \(H\) separately and combines their outputs using schema and entity mappings \cite{tkde}. By operating directly on the factorized data, this approach significantly improves the efficiency of model training while maintaining predictive accuracy. We will detail the example in Sec.~
\ref{s:operations}.


Despite widespread use of LA representations in current factorization learning research \cite{\acdc,chen2017towards,MorpheusFI}, few studies have attempted to leverage the capabilities of LA-friendly GPUs. GPUs are extensively used in modern ML applications, significantly accelerating both training and inference speeds for deep learning \cite{pytorch} and traditional statistical ML models \cite{wenthundersvm18,ke2017lightgbm}. This oversight leads to missed opportunities for significant speedups on GPUs. While specific algorithms \cite{gpu1} are designed to efficiently train models on contemporary hardware, achieving this often requires manually crafted LA rewriting to ensure compatibility.

Moreover, factorized learning does not consistently outperform training over materialized data. Heuristic rules \cite{MorpheusFI,chen2017towards} based on data characteristics frequently overlook intricate interactions among data characteristics, training algorithms, and hardware environments. This shortcoming is particularly evident in batch ML training workloads, such as those used for dataset selection \cite{chepurkoARDAAutomaticRelational2020, galhotra2023metam} and hyperparameter tuning, where overly strict thresholds may restrict potential speed gains, while overly relaxed thresholds could degrade performance. Consequently, a precise cost estimator is essential to enhance the practical usability of factorized learning in ML training workloads.
\begin{figure*}[t]
    \centering
    \includegraphics[width=0.95\linewidth]{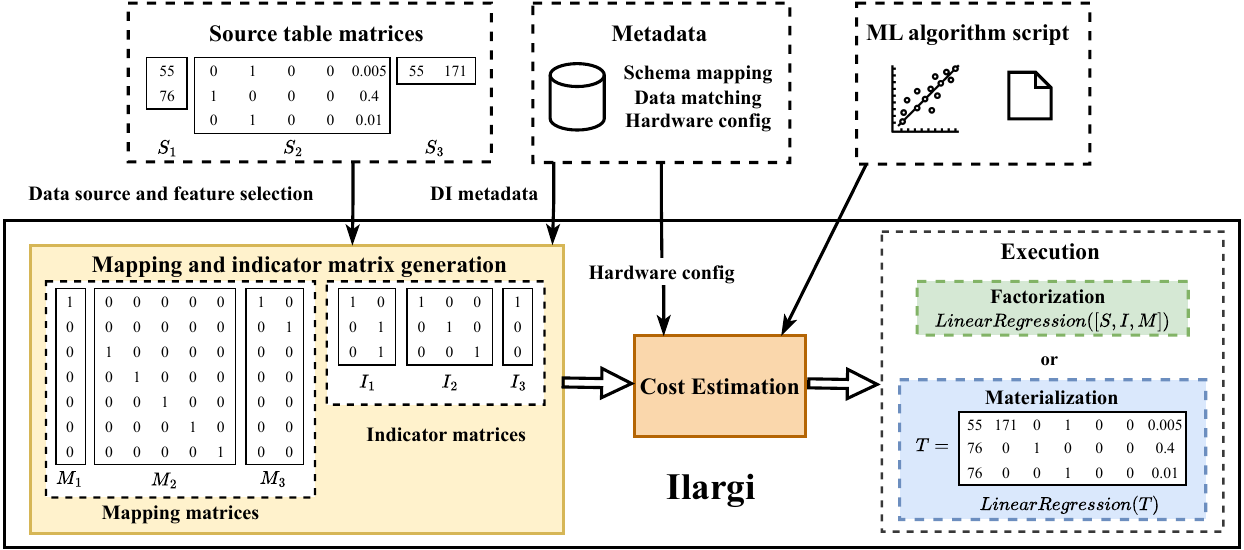}
    \caption{Ilargi's workflow.} 
    \label{fig:amalur}
    \vspace{-5mm}
\end{figure*}

To bridge these research gaps, we present \emph{Ilargi}\footnote{Complete technical details, code, and datasets are available at \url{https://github.com/amademicnoboday12/Ilargi/}}, an efficient factorized learning framework for training ML models on both CPU and GPU platforms. The novelty of this work lies in its GPU-compatible representation and a learning-based cost estimator. First, we employs a matrix representation of data integration (DI) metadata, which unifies and streamlines both DI and machine learning operations. This unified representation enables highly efficient factorized model training on both CPUs and GPUs. Furthermore, we propose a cost estimator within \emph{Ilargi} that considers memory access behavior to suggest the optimal training strategy and avoid performance degradation due to the misuse of factorized learning. These innovations collectively improve the practical usability of factorized learning. We summarize the contributions of this work as follows:

\begin{itemize}
\vspace{-1mm}
\item \emph{Ilargi} models data integration (DI) and ML training as a sequence of linear algebra (LA) operations (Sec. \ref{sec:system}), enabling \textbf{automated} LA rewriting to transform learning algorithms into factorized forms. This approach is fully \textbf{GPU-compatible}, significantly boosting training speed across various ML tasks. Experiments (Sec. \ref{ssec:simpleDI}) show up to 8.9x speedups on GPUs over CPU implementations.

\item \emph{Ilargi} introduces a \textbf{learning-based cost estimator} (Sec. \ref{sec:cost}) to predict the optimal training method—factorized or materialized—based on data and hardware characteristics. The estimator achieves 97.1\% accuracy, boosting batch training performance by over 20\% compared to materialized data training (Sec. \ref{ex:estimator}).

\vspace{-1mm}
\end{itemize}



\vspace{-3mm}
\section{Ilargi: ML Over Disparate Sources}
\label{sec:system}


\vspace{-3mm}
As illustrated in Fig.~\ref{fig:amalur}, \emph{Ilargi} first converts the inputs, DI metadata and source tables, into sparse matrices, storing them in a compressed format. This transformation facilitates a GPU-compatible representation of schema mappings and data matching  (Sec.~\ref{sec:matrixGen}). \emph{Ilargi} then performs LA rewriting to both ML model training algorithms and data integration tasks, reconfiguring the process into a sequence of linear algebra operations (Sec.~\ref{s:operations}). The cost estimator subsequently predicts the optimal training method—either factorization or materialization—based on data characteristics, algorithmic complexity, and the hardware environment (Sec.~\ref{sec:cost}). Finally, \emph{Ilargi} trains the model with the optimal training strategy recommended by the cost estimator.

In this section, we introduce how \emph{Ilargi} utilizes matrix-represented data integration metadata to enable efficient factorized model training on both GPUs and CPUs.


\vspace{-3mm}
\subsection{DI Metadata as Matrices} 
\label{sec:matrixGen}
\vspace{-2mm}
Preparing disparate training data for ML models requires understanding the relationships between the data sources and bridging them, which is achieved primarily through data integration tasks. The relationships are normally described by following metadata: \cite{doan2012principles}: $i)$ mappings between different source schemata, i.e., schema matching and mapping \cite{rahm2001survey, fagin2009clio} and $ii)$ linkages between data instances, i.e., data matching (also known as record linkage or entity resolution)  \cite{brizan2006survey}. 
We refer to such vital information derived from data integration process as \emph{data integration metadata} (\textit{DI metadata}).

Schema mappings are traditionally formalized as first-order logic~\cite{fagin2009}, and represented and executed using query languages such as SQL~\cite{yan2001data} or XQuery~\cite{fagin2009clio}. To accelerate both data integration and machine learning operations using GPUs, it is critical to find a unified representation that is highly compatible with GPU architectures.
In Ilargi, we employ a matrix-based representation for schema mappings, which we refer to as \emph{mapping matrices}, denoted as $\mathbf{M}$. Given a source table $S_k$ and target table $T$, a mapping matrix $M_k \in \mathbf{M}$ has the size of  $c_T \times c_{k}$, and specifies how the columns of $S_k$ are mapped to the columns of $T$. 

\begin{definition}[\textit{Mapping matrix}] Mapping matrices between source tables $S_1, S_2, ..., S_n$ and target table $T$ are a set of binary matrices $\mathbf{M}= \{M_1, ..., M_n\}$. $M_k \ (k\in [1, n])$ is a matrix with the shape $c_{T} \times c_{k}$, where  

\small{
		\begin{align*}
			\begin{split} 
M_k[i,j] =  \begin{cases}1, &\text{if j-th column of $S_k$ mapped to i-th column of } T\\0, & \text{otherwise}\end{cases}
			\end{split}
		\end{align*}
	}
	\label{def:mm}
 
\end{definition}

\vspace{-2mm}
Intuitively, in $M_k[i,j]$ the vertical coordinate $i$ represents the target table column while the horizontal coordinate $j$ represents the mapped source table column.  A value of $1$ in $M_k$ specifies the existence of  column correspondences between  $S_k$ and $T$, while the value $0$ shows that the current target table attribute has no corresponding column in $S_k$. Fig.~\ref{fig:amalur} shows the mapping matrices $M_1$, $M_2$, $M_3$ for source tables $S_1$, $S_2$, $S_3$, respectively. 



We use the \emph{indicator matrix} \cite{chen2017towards} 
to preserve the row matching between each source table $S_k$ and the target table $T$. 
An indicator matrix $I_k$ of size $c_T \times c_{k}$ describes how the rows of source table $S_k$ map to the rows of target table $T$, as in Fig.~\ref{fig:amalur}. 
 
\begin{definition}[\textit{Indicator matrix} \cite{chen2017towards}] Indicator matrices between source tables $S_1, S_2, ..., S_n$ and target table $T$ are a set of row vectors $\mathbf{I}= \{I_1, ..., I_n\}$. $I_k \ (k\in [1, n])$ is a row vector  of size $r_{T}$, where 
\small{
		\begin{align*}
			\begin{split} 
					I_k[i,j] =  \begin{cases}
$1$ & \text{if $j$-th row of $S_k$ mapped to $i$-th row of $T$} \\
$0$ & \text{otherwise}
\end{cases}
			\end{split}
		\end{align*}
	\label{def:cim}
}
\end{definition}


\begin{table}[t]
\caption{Notations used in the paper.}
\label{tab:notations}
\vspace{1mm}
\centering
\footnotesize
\begin{tabular}{|l|l|} \hline
\textbf{Symbol} & \textbf{Description} \\ \hline
$S_k$ & The $k$-th source table \\ \hline
$T$ & Target table \\ \hline
$I_k$ & The indicator matrix for $S_k$\\ \hline
$M_k$ & The mapping matrix for $S_k$ \\ \hline
$j_T$          & The join type of $T$ (inner/left/outer/union) \\ \hline
$r_k/r_T$      & Number of rows in $S_k/T$ \\ \hline
$c_k/c_T$      & Number of columns in $S_k/T$ \\ \hline
$m_k/m_T/m_X/m_w$      & Number of nonzero elements in $S_k/T/X/w$ \\ \hline
\end{tabular}
\vspace{-4mm}
\end{table}

\vspace{-5mm}
\subsection{Rewriting Rules for Factorization }
\label{s:operations}
With data and metadata represented in matrices, next, we explain how to rewrite a linear algebra over a target schema to linear algebra over source schemas. 
Here we use the example of LA operator \emph{left matrix multiplication (LMM)}. 
The full set of LA rewrite rules based on mapping and indicator matrices is in our technical report \cite{tech}.

\para{Left Matrix Multiplication} Given a matrix $X$ with the size $c_T \times c_X$, LMM of T and X is denoted as $TX$.  The  LMM result of our target table matrix $T$ and another matrix $X$ is a matrix of size $r_T \times c_X$. Our rewrite of LMM goes as follows.
\[
TX \rightarrow  I_1 S_1 M_1^T X + ... + I_n S_n M_n^T X
\]
We first compute the local result  $I_{k} S_k M_k^{T}X\ (k\in [1, n])$ for each source table, 
then assemble them for the final results.

%


To showcase the effectiveness of matrix-represented DI metadata for factorized learning, we use Linear Regression as an example. In Alg.~\ref{alg:linear-regression}, we identify two key operators: matrix transpose and left matrix-matrix multiplication. By applying our rewriting strategies, these operations are pushed directly to the disparate data sources, enabling local computation and eliminating the need for data centralization, thus improving efficiency and scalability.

\begin{algorithm}[t!]
  \caption[Linear regression]{Linear regression using Gradient Descent
    ~\cite{chen2017towards}.}\label{alg:linear-regression}
  \begin{algorithmic}
    \Require $X, y , w, \gamma$
    \For{$i \in 1:n$} 
     \State $w = w - \gamma (\text{{$X^T$}}((\text{{$X w$}}) - y))$
    \EndFor
  \end{algorithmic}
\end{algorithm}
\vspace{-4mm}
\section{Cost Estimation in Ilargi}
\label{sec:cost}
\vspace{-2mm}
Reducing ML model training time through factorization is not always guaranteed. When the target table has low redundancy compared to source tables, factorization may require more linear algebra computations, leading to longer training times than materializing the target table \cite{chen2017towards}. Existing studies \cite{chen2017towards, MorpheusFI} use empirical cost models to set sparsity-based thresholds for selecting between factorization and materialization, yet these thresholds vary with hardware.  However, in environments with both CPUs and GPUs, our experiments (Sec. \ref{sec:eva}) show that target table sparsity does not consistently correlate with factorization speedups, and identical sparsity levels can yield different speedup outcomes on different hardware platforms.

This discovery reveals the intricacy of selecting the optimal training method between factorization and materialization. Consequently, it is essential to develop a cost model that considers both algorithmic characteristics and hardware properties. To build this cost model, we begin with a complexity analysis of model training. This analysis enables a comparative evaluation of the complexities associated with materialization and factorization.


\vspace{-4mm}
\subsection{Complexity Analysis of ML Model Training Algorithms}
\label{s:complexity}
\vspace{-2mm}
We compare the computational complexity of linear algebra (LA) operations for both materialization and factorization. For materialization, the cost is based on performing LA operations directly on the materialized table. In contrast, factorization considers the operations on each source table individually.

\begin{table}[t]
\caption{Computations complexity comparison of common LA operators.}
\label{tab:complexity}
\vspace{1mm}
\centering
  \small
\begin{tabular}{l|c|c}
\toprule
\textbf{Operation} & \textbf{Materialization}                  & \textbf{Factorization}                  \\ \midrule
$T \oslash x$  & \multirow{4}{*}{$\displaystyle m_T$} & \multirow{4}{*}{$\displaystyle \sum_{k=1}^n m_k$} \\
 $f(T)$                  &                                    &                                      \\
$\text{rowSum}(T)$       &                                    &                                      \\
$\text{colSum}(T)$       &                                    &                                      \\ \hline

$TX$               & $\displaystyle c_X \cdot m_T + r_T \cdot m_X$              & $\displaystyle \sum_{k=1}^n c_X \cdot m_k + r_k \cdot m_X$            \\
$XT$               & $\displaystyle  r_X \cdot m_T + c_T \cdot m_X $              & $\displaystyle \sum_{k=1}^n   r_X \cdot m_k + c_k \cdot m_X$      \\
\bottomrule
\end{tabular}
\vspace{-5mm}
\end{table}

Tab.~\ref{tab:complexity} summarizes the comparison. Ilargi stores matrices in sparse format, so the complexity of element-wise operations, rowSum, and colSum depends on the number of non-zero elements.

For sparse matrix multiplication, given two matrices \( T \) and \( X \), the complexity of \( TX \) is \( O(c_X \cdot m_T + r_T \cdot m_X) \), where \( m_T \) and \( m_X \) are the non-zero elements of \( T \) and \( X \) \cite{horowitz1982fundamentals}. Since our mapping and indicator matrices are sparse, we calculate matrix operation costs accordingly. With the rewriting rule (Sec.~\ref{s:operations}), the complexity of factorized multiplication is \( \sum_{k=1}^n (c_X \cdot m_k + r_k \cdot m_X) \).

The complexity of ML models depends on the specific LA operators used. We detail linear regression analysis here, with further models discussed in our technical report \cite{tech}.

\para{Linear regression} First, we analyze the complexity of linear regression in Algorithm \ref{alg:linear-regression} in the materialized case, which is dominated by two matrix multiplication operations, i.e., $ T^T$ $(T w)$. 
For the first matrix multiplication $T w$: 
we denote the shape of weights vector $w$ as $c_T \times 1$. Following the general assumption that $w$  is dense, we calculate $m_w$, the number of nonzero elements in $w$, with the equation $m_w = r_w \times c_w$. 
We denote the matrix multiplication result of $T w$ as $X$, which is an intermediate result
of linear regression algorithm. The size of $X$ is $r_T \times 1$, since $r_X= r_T\  and \ c_X =1$. 
Similarly, we calculate $m_X$, the number of nonzero elements in $X$, with the equation $m_X = r_X \times c_X$.  
Now, we define the complexity of linear regression in the materialized case based on $T w$ and $T^TX$.
\[
O_\text{materialized}(T) = \underbrace{c_w \cdot m_T + r_T \cdot m_w}_{Tw} + \underbrace{m_T + c_T \cdot m_X}_{T^TX}
\]

Next, we define the complexity of the factorized case. 
\[
\begin{aligned}
O_\text{factorized}(T) = & \underbrace{\sum_{k=1}^n (c_w \cdot m_k + r_k \cdot m_w)}_{Tw} + 
 \underbrace{\sum_{k=1}^n (m_k + c_k \cdot m_X)}_{T^TX}
\end{aligned}
\]

\para{Complexity ratio}
We define a variable \emph{complexity ratio} to indicate whether materialization or factorization leads to more computing. 
The complexity ratio is measured as the ratio of the materialization complexity divided by the factorization complexity.
\begin{equation}
\label{cr_value}
     \text{complexity ratio} = \frac{O_\text{materialization}(T)}{O_\text{factorization}(T)}
\end{equation}

\vspace{-2mm}
\subsection{The Hardware Factor} 
\label{sec:3rdfac}
\vspace{-2mm}
In our evaluation (detailed in Sec. \ref{fig:eval_f_m}), we found that materialized training can outperform factorized learning, despite the higher redundancy. Additionally, GPU speedups are inconsistent. These discrepancies go beyond computational complexity, highlighting the influence of hardware properties like memory bandwidth and parallelism. Traditional complexity analysis often neglects the I/O overhead from data movement, especially in multi-epoch ML training where I/O costs accumulate. To address this, we introduce a learned cost estimator that considers data characteristics, algorithm complexity, and hardware features to better determine the optimal training method.

\vspace{-3mm}
\subsection{ML-based Cost Estimator}
\label{s:estimator}
\vspace{-2mm}

Our cost estimator determines whether factorization or materialization is the more efficient training method, considering data characteristics, computational complexity, and hardware features. This choice is especially impactful when training multiple models, such as in parameter tuning, where small speed gains can lead to significant time savings.

Analytical performance models \cite{hpc2,hpc3} from high-performance computing are limited in flexibility for diverse ML models and require extensive micro-benchmark data, making a black-box estimator \cite{blackbox1} a practical alternative. This type of estimator uses a statistical model to make a binary decision, abstracting details of hardware and algorithms.

\para{Ilargi's Tree Boosting Estimator.} Tree boosting is known for its explainability and speed, making it popular in cost estimation tasks \cite{tvm,halide}. We use XGBoost for its ability to capture non-linear relationships, with a focus on identifying key features that influence the choice between materialization and factorization.

\para{Hardware Features.} To ensure portability, our estimator emphasizes macro-level hardware features, such as memory bandwidth and parallelism, over micro-level specifics like cache speed. Parallelism, particularly the number of threads used, is included due to its impact on factorized learning speedups. Combined with memory read/write costs, these features enable effective cost estimation for both methods (Sec.~\ref{ex:estimator}).

\begin{figure}[t]
    \centering
    \includegraphics[width=0.8\linewidth]{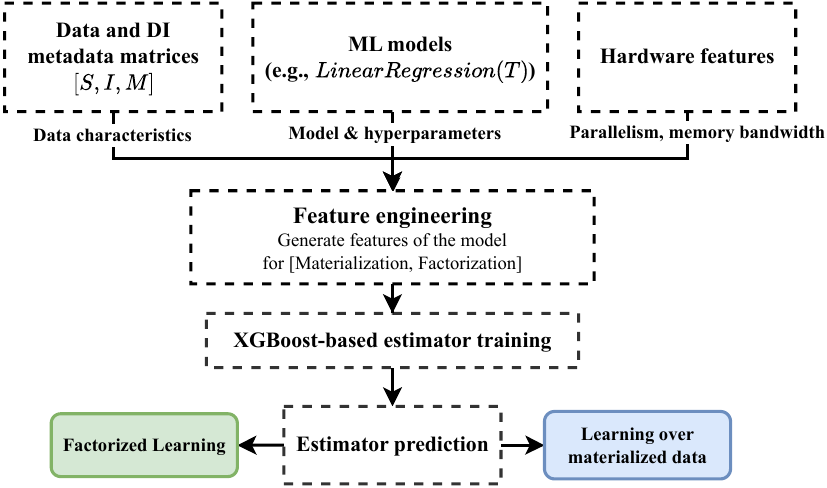}
    \caption{Workflow of the estimator.}
    \label{fig:cost_estimation_amalur}
    \vspace{-5mm}
\end{figure}

\para{Cost estimation pipeline}\label{s:design}
The cost estimator uses three input groups: \texttt{i)} data (source matrices, mapping, and indicator matrices), \texttt{ii)} ML algorithm (operators and user-defined hyperparameters), and \texttt{iii)} hardware (e.g., parallelism, memory bandwidth). Fig. \ref{fig:cost_estimation_amalur} shows its workflow using linear regression as an example. The estimator\footnote{A full list of 33 features is detailed in our technical report \cite{tech}.} computes the complexity ratio (Sec.~\ref{s:operations}) and the theoretical memory I/O for each operator in the ML algorithm. It normalizes these features based on parallelism and memory bandwidth. The output is binary: If True, Ilargi uses factorization; otherwise, it materializes the data before training.

\section{Evaluation}
\label{sec:eva}
\vspace{-2mm}
We start this section with an overview of our experimental setup, followed by a detailed performance evaluation of machine learning workloads employing both factorization and materialization techniques on CPUs and GPUs. This evaluation specifically examines the computing performance of factorized ML model training on these platforms and explores the factors that influence the decision-making process between factorization and materialization. Building on these insights, we further evaluate the accuracy of our proposed cost estimator in predicting the optimal strategy, utilizing both real and synthetic datasets. Our results demonstrate that the cost estimator can reduce the time consumption of real-world model training workloads by more than 20\%.

\vspace{-3mm}
\subsection{Experiment Setup}
\label{s:setup}

\vspace{-1mm}
\para{Synthetic Datasets}
\label{s:setup_synthetic} 
We built a data generator\footnote{\url{https://github.com/amademicnoboday12/Ilargi/tree/main/src/data_generator}} to create synthetic datasets with diverse data characteristics. It generates source table pairs that join to form a target table based on specified parameters. Using the ranges in Tab.~\ref{tab:synth_data_char}, we produced 1800 datasets to capture the relationship between data features and training performance. Additionally, the results from these datasets serve as training data for our cost estimator.

\begin{table}[t]
\caption{Parameters used for synthetic dataset generation.}
\label{tab:synth_data_char}
\vspace{1mm}
\centering
\footnotesize
\begin{tabular}{|l|l|l|}
\hline
\textbf{Symbols}           & \textbf{Ranges}             & \textbf{Description}                                                                                                            \\ \hline
$r_T$, $r_{S_k}$   & 100k, 500k, 1M     & \begin{tabular}[c]{@{}l@{}}Number of rows in target table $T$ or source \\table $S_k$ \end{tabular}                                        \\ \hline
$c_T$, $c_{S_k}$   & {[}10, 50{]}       & \begin{tabular}[c]{@{}l@{}}Number of columns in target  table $T$ or source \\table $S_k$\end{tabular}                                    \\ \hline
$p$              & {[}0, 0.9{]}       & \begin{tabular}[c]{@{}l@{}}Sparsity of the target table\end{tabular}                             \\ \hline
$\rho_c(T)$      & {[}0.1, 1{]}       & \begin{tabular}[c]{@{}l@{}} The percentage of number of columns in source \\tables  w.r.t target table\end{tabular} \\ \hline
$j$              & \begin{tabular}[c]{@{}l@{}}Inner, left, or \\outer joins, union\end{tabular}     & Join types                                                                                                             \\ \hline
\end{tabular}
\vspace{-4mm}
\end{table}


\para{Real-World Datasets}
To validate the estimator in real scenarios, we used the Hamlet datasets \cite{2016-hamlet-sigmod}, commonly referenced in related works \cite{MorpheusFI,chen2017towards,orion_learning_gen_lin_models}. The seven datasets in Hamlet simulate ML workflows, originally designed for inner join scenarios but adapted for other join types as detailed in Tab.~\ref{tab:6-hamlet-characteristics}.

\begin{table}[t]
  \centering
    \caption[Hamlet dataset characteristics]{Hamlet dataset characteristics. $r$ and $c$ indicate the number of rows and columns. Subscripts denote which table characteristics belong to. }
  \label{tab:6-hamlet-characteristics}
  \vspace{1mm}
  \footnotesize
  \begin{tabular}{|l|l|l|l|l|l|l|l|l|l|l|}
  \hline
     & $r_T$ & $c_T$ & $r_{S_1}$ & $r_{S_2}$ & $r_{S_3}$ & $r_{S_4}$ & $c_{S_1}$ & $c_{S_2}$ & $c_{S_3}$ & $c_{S_4}$ \\ \midrule \midrule
    Expedia & $942$K  & $52.3$K & $942$K  & $11.9$K & $37$K   & -    & $27$    & $12$K   & $40.2$K & -      \\
    Flight  & $66.5$K & $13.7$K & $66.5$K & $540$   & $3.17$K & $3.17$K & $20$    & $718$   & $6.46$K & $6.47$K \\
    Lastfm  & $344$K  & $55.3$K & $5$K    & $50$K   & -       & -    & $5.02$K & $50.2$K & -      & -       \\
    Movie   & $1$M    & $13.3$K & $6.04$K & $3.71$K & -       & -    & $9.51$K & $3.84$K & -      & -       \\
    Yelp    & $216$K  & $55.6$K & $11.5$K & $43.9$K & -       & -    & $11.7$K & $43.9$K & -      & -       \\
    \bottomrule
  \end{tabular}
\vspace{-2mm}
\end{table}

\para{TPC-DI Benchmark} 
For large-scale evaluation, we used the TPC-DI benchmark~\cite{tpcdi}, involving four datasets across three sources. The fact table, \emph{Trade}, is joined with key attributes to produce a target table with 27 features. Tab.~\ref{tab:tpc_data} outlines table cardinalities across various scale factors, highlighting data characteristics in different scenarios. 

\begin{table}[t]
\caption{Data sizes of realistic data integrations scenario based on TPC-DI benchmark. The table shows the number of rows in sources and  target table w.r.t varying scale factors.}
\label{tab:tpc_data}
\vspace{1mm}
\centering
\footnotesize
\begin{tabular}{|c|l|l|ll|l|}
\hline
              & \textbf{Source 1} & \textbf{Source 2}  & \multicolumn{2}{c|}{\textbf{Source 3}}         &    \textbf{Target}      \\ \hline
\textbf{Scale factor} & Trade   & Customer & \multicolumn{1}{l|}{Stock} & Reports &  -   \\ \hline
3             & 391.1K & 4.7K    & \multicolumn{1}{l|}{2.6K} & 98.7K  & 528.8K   \\ \hline
7             & 911.6K & 108K   & \multicolumn{1}{l|}{5.8K} & 297.7K & 1.3M \\ \hline
\end{tabular}
\vspace{-4mm}
\end{table}


\para{Hardware and software} We run experiments with 16, and 32 cores of AMD EPYC 7H12 CPU, and Nvidia A40 GPU. To enable parallelized sparse matrix multiplication on both CPU and GPU, we choose MKL and CuBLAS as LA library respectively. The number of model training iterations is 200. 

\vspace{-2mm}
\subsection{Performance Evaluation with Synthetic Data}
\label{ssec:simpleDI}
\vspace{-2mm}
In this section, we conduct a performance comparison between factorized learning and learning over materialization on both CPUs and GPUs. The purpose of this comparison is to demonstrate the significant acceleration that GPUs can provide for factorized learning, as well as the performance of factorized learning across various data characteristics.

\vspace{-3mm}
\subsubsection{Performance of factorization and materialization}
\begin{figure}[t]
    \centering
    \includegraphics[width=0.8\linewidth]{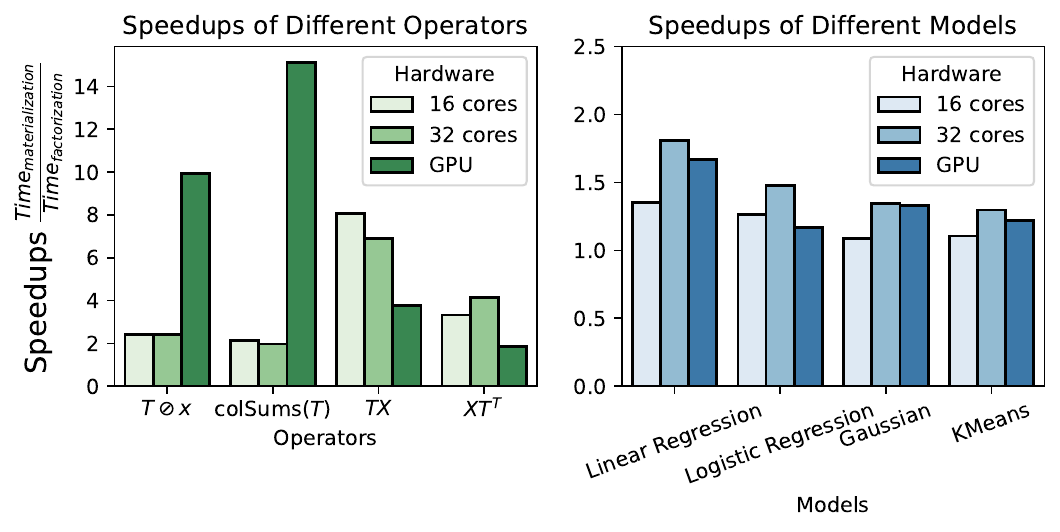}
        \vspace{-1.5mm}
    \caption{
    Speedups (\(\frac{\text{Time}_{materialization}}{\text{Time}_{factorization}} \)) of LA operators and model training w.r.t varying input parameters. \emph{Here we focus on the cases that factorization performs faster than materialization.}
    }
    \label{fig:eval_f_m}
    \vspace{-4mm}
\end{figure}

Fig.~\ref{fig:eval_f_m} shows the speedups of factorization over materialization (\(\frac{T_{materialization}}{T_{factorization}}\)) across three hardware setups. Factorized scalar operators achieve greater speedups on GPUs compared to CPUs, while matrix operators show smaller gains. On CPUs, matrix operators outperform scalar operators, but this trend reverses on GPUs. This discrepancy arises because GPU-based matrix operations are often limited by memory bandwidth, as discussed in Section~\ref{sec:3rdfac}. The high parallelism of GPUs generally benefits compute-bound tasks more than memory-bound operators.

For model training, factorization still offers speed improvements, albeit more modestly. This is due to the frequent read/write operations for intermediate results in training algorithms, which diminishes speedups, especially on GPUs where memory bandwidth constraints often dominate.

\vspace{-3mm}
\subsubsection{Speedups regarding sparsity and complexity ratio}
\begin{figure}[t]
    \centering
    \includegraphics[width=\linewidth]{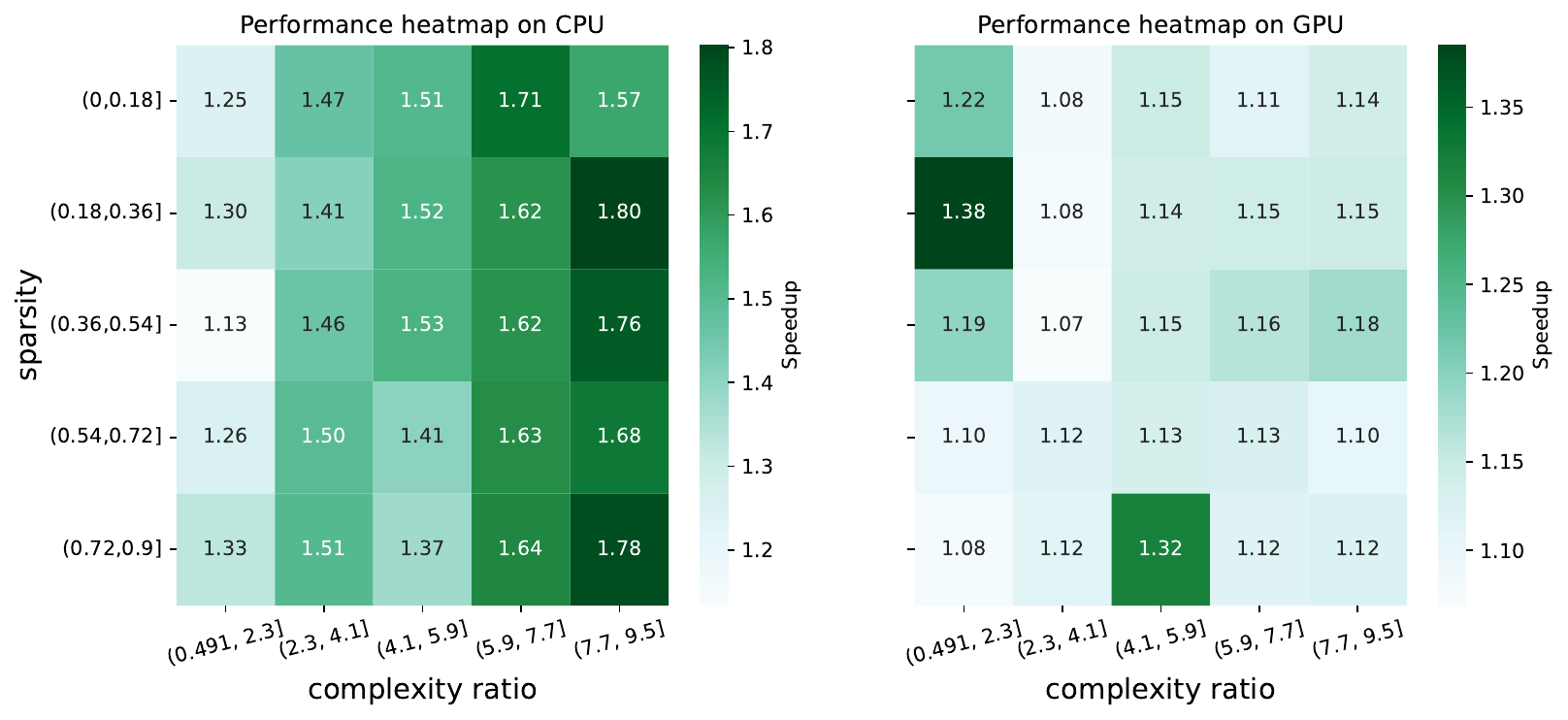}
    \caption{
        Speedups (\(\frac{\text{Time}_{materialization}}{\text{Time}_{factorization}} \)) w.r.t target table sparsity and complexity ratio on CPUs and GPUs. On CPUs, speedups increase when complexity ratio gets larger. No observable trend on GPUs. 
    }
    \label{fig:heatmap_cpu_gpu}
    \vspace{-4mm}
\end{figure}

Fig.~\ref{fig:heatmap_cpu_gpu} illustrates how speedups vary with varying sparsity and complexity ratio on CPUs and GPUs. The speedups presented in the figure represent the average benefit when factorization proves more advantageous. From the CPU results, an increase in speedups is observed as the complexity ratio grows. Within certain intervals of complexity ratio, speedups increase as sparsity decreases. However, sparsity alone is not a reliable estimator across all complexity ratios.

The results on GPUs, in contrast, show no observable trend, suggesting that empirical threshold-based estimators may not function effectively on GPUs. This observation underscores the necessity for a learned estimator capable of integrating features of both hardware configurations and data characteristics.


\mybox{\textbf{Takeaways:} 
The significant variability in performance comparison between factorization and materialization, highlights \textbf{the need for a cost estimator}. 
}

\vspace{-1mm}
\subsection{Effectiveness of Our Estimator}
\label{ex:estimator}
\vspace{-2mm}
This section evaluates the tree boosting cost estimator as outlined in Section \ref{s:estimator}. Our analysis focuses not only on the predictive accuracy of the estimator but also on its performance enhancement in batch model training scenarios. 


\para{Training and test data}
We extract features described in Sec.~\ref{s:estimator} from the evaluation conducted on synthetic data. This process generated 7200 pairs (1800*4), each labeled to indicate whether factorization is faster than materialization (True) or not (False). The dataset, containing these features and labels, is then randomly divided into training (80\%) and test (20\%) sets.

\para{Evaluation metrics} 
We evaluate our estimator model using accuracy and F1-score, and compare its performance against other baseline estimators. Accuracy serves to gauge the effectiveness of our cost estimator in accurately recommending the faster training method—either factorization or materialization. Furthermore, we measure the end-to-end speedup of batch model training tasks when decisions between factorization and materialization are guided by our cost estimator. 

\para{Four baselines for comparison} 
\texttt{i)} To confirm the significance of hardware features, 
we train another tree boosting model using the same dataset, but \emph{without hardware features}. 
\texttt{ii)} The \emph{tuple ratio and feature ratio} (TR \& FR), metrics used in Morpheus \cite{chen2017towards}, quantify the target table's redundancy relative to the joinable tables. Morpheus \cite{chen2017towards} suggests a threshold of 5 for tuple ratio and 1 for feature ratio.
 \texttt{iii)} Finally, we consider the heuristic rule proposed in \cite{MorpheusFI}.
\begin{table}[t]
\caption{Accuracy, F1-score and Overall Speedups of estimators tested on CPUs and GPUs. Our cost estimator shows superior predictive quality and effectiveness over synthetic test data.}
\label{tab:overall_accuracy}
\small
\centering
\begin{tabular}{|l|l|l|l|}
\hline
\textbf{Cost estimators}                        & \textbf{Accuracy}       & \textbf{F1-score}  & \textbf{Overall Speedups} \\  \hline
\multicolumn{4}{|c|}{CPU results} \\ \midrule
Tree Boosting (ours)          & \textbf{0.971} & \textbf{0.936} & \textbf{1.24} \\ \hline
    Tree Boosting w/o hardware        &0.712 & 0.560 & 1.19\\ \hline
TR \& FR \cite{chen2017towards}            & 0.490          & 0.426 & 1.16\\ \hline
MorpheousFI             & 0.864          & 0.601        & 1.22  \\ \hline
\hline
\multicolumn{4}{|c|}{GPU results} \\ \midrule
Tree Boosting (ours)          & \textbf{0.899} & \textbf{0.821} & \textbf{1.15} \\ \hline
    Tree Boosting w/o hardware        &0.523 & 0.443 & 1.09\\ \hline
TR \& FR \cite{chen2017towards}            & 0.244          & 0.362 & 0.86\\ \hline
MorpheousFI             & 0.497          & 0.477        & 0.96  \\ \hline
\end{tabular}
\vspace{-4mm}
\end{table}

\vspace{-2mm}
\subsubsection{Results on synthetic data} 
Tab.~\ref{tab:overall_accuracy} shows that our tree boosting model outperforms others with a 97.1\% accuracy and high F1-score, confirming its strong precision-recall balance. Excluding hardware features significantly reduces both accuracy and F1-score, supporting the importance of hardware information in choosing between factorization and materialization. The two threshold-based estimators perform worse in both metrics.

Beyond quality metrics, our estimator achieves a 1.24x speedup over training exclusively with materialization, outperforming all baselines. On GPUs, however, the effectiveness of all estimators declines, with threshold-based models performing worse than default materialization. This highlights the sensitivity of empirical thresholds to hardware variations, as discussed in Sec.~\ref{sec:3rdfac}, and demonstrates the limitations of threshold-based methods in cross-platform tasks.


\vspace{-5mm}
\subsubsection{Results on real-world data}
We further evaluated our estimator using the \textit{Project Hamlet} and \textit{TPC-DI} datasets to test its practical usability in realistic ML tasks, including multiple model training and hyperparameter tuning.

Tab.~\ref{tab:overall_estimator} shows that the speedups on both CPU and GPU platforms are consistent with those observed on synthetic data, with our tree boosting estimator consistently performing best on unseen data.
\begin{table}[t!]
\caption{Average speedups (\(\frac{\text{Time}_{materialization}}{\text{Time}_{factorization}} \)) of all models across all real-world data. Our tree boosting estimator can consistently accelerate batch model training workload on both CPUs and GPUs.}
\label{tab:overall_estimator}
\centering
\small
\begin{tabular}{|l|l|l|}
\hline
\textbf{Cost estimators} & \textbf{Speedups (CPU)} & \textbf{Speedups (GPU)} \\ \hline
Tree boosting (ours) & \textbf{1.21} & \textbf{1.16} \\ \hline
MorpheousFI & 1.18 & 0.88 \\ \hline
TR\&FR & 1.09 & 072 \\ \hline
\end{tabular}%
\vspace{-4mm}
\end{table}

Tab.~\ref{tab:effect-cpu} provides a breakdown of effectiveness across models and datasets on CPUs. While the estimator generally makes accurate decisions, its performance varies with different datasets. For denser datasets like \textit{flight} and \textit{lastfm}, incorrect decisions led to slower training compared to materialization. Still, when training a single model across various datasets, speedups reached up to 1.29x.

On GPUs (Tab.~\ref{tab:effect-gpu}), the results mirror the CPU findings but show slightly more mis-predictions. Speedups are generally lower on GPUs, consistent with synthetic data results. Predicting accurately for dense datasets like \textit{flight} and \textit{lastfm} remains challenging, as this high density is atypical for most star schema datasets.

\mybox{\textbf{Takeaways:}  Our tree boosting cost estimator, enhanced with hardware features, outperforms the state-of-the-art estimator, showing consistent effectiveness across diverse hardware platforms.}

\begin{table}[t!]
\caption{Detailed speedups(\(\frac{\text{Time}_{materialization}}{\text{Time}_{factorization}} \)) w.r.t different models and datasets on CPUs (32 cores). Our tree boosting estimator makes correct prediction in most cases but performs worse on \textit{flight} and \textit{lastfm}.}
\label{tab:effect-cpu}
\vspace{1mm}
\small
\centering
\begin{tabular}{p{25mm}|cccc|c}
\tikz[diag text/.style={inner sep=0pt, font=\footnotesize},
      shorten/.style={shorten <=#1,shorten >=#1}]{%
        \node[below left, diag text] (def) {Datasets};
        \node[above right=2pt, diag text] (abc) {Models};
        \draw[shorten=4pt, very thin] (def.north west|-abc.north west) -- (def.south east-|abc.south east);}
& \cellcolor{white} G.NMF & \cellcolor{white} KMeans & \cellcolor{white} Lin.Reg & \cellcolor{white} Log.Reg & \cellcolor[HTML]{EFEFEF} Speedups/dataset \\ \hline
\rowcolor{green!5} 
\cellcolor{white} TPC-DI sf=3                                                   & 1.33                             & \cellcolor{red!10}1      & 1.12                            & 1.33                            & 1.10                                                                                    \\ \hline
\rowcolor{green!5} 
\cellcolor{white} TPC-DI sf=7                                                   & 1.23                             & \cellcolor{red!10}1      & 1.04                            & 1.20                            & 1.09                                                                                    \\ \hline
\rowcolor{green!5} 
\cellcolor{white} expedia                                                & 1.24                             & 1.56                           & \cellcolor{red!10}1       & 1.07                            & 1.17                                                                                    \\ \hline
\rowcolor{red!10}  
\cellcolor{white} flight                                                 & 0.57                             & 0.7                            & \cellcolor{green!5}1       & 0.49                            & 0.74                                                                                    \\ \hline
\rowcolor{green!5} 
\cellcolor{white} lastfm                                                 & 1                                & \cellcolor{red!10}0.9    & 1                               & 1                               & \cellcolor{red!10}0.96                                                            \\ \hline
\rowcolor{green!5} 
\cellcolor{white} movie                                                  & 1.8                              & 2.35                           & 2.32                            & 1.88                            & 2.19                                                                                    \\ \hline
\rowcolor{green!5} 
\cellcolor{white} yelp                                                   & 1.10                             & 1.24                           & 1.51                            & \cellcolor{red!10}1       & 1.28                                                                                    \\ \hline\hline
\cellcolor[HTML]{EFEFEF}\begin{tabular}[c]{@{}l@{}}Speedups/model \end{tabular} & \cellcolor{green!5}1.28     & \cellcolor{green!5}1.21   & \cellcolor{green!5}1.29    & \cellcolor{green!5}1.18    & \#                                                                                        \\ 
\end{tabular}
\vspace{-3mm}
\end{table}

\begin{table}[t]
\caption{Detailed speedups(\(\frac{\text{Time}_{materialization}}{\text{Time}_{factorization}} \)) w.r.t different models and datasets on GPU. Our tree boosting estimator makes more wrong predictions then that on CPUs but still achieves overall speedups in most cases.}
\label{tab:effect-gpu}
\vspace{1mm}
\small
\centering
\begin{tabular}{p{25mm}|cccc|c}
\tikz[diag text/.style={inner sep=0pt, font=\footnotesize},
      shorten/.style={shorten <=#1,shorten >=#1}]{%
        \node[below left, diag text] (def) {Datasets};
        \node[above right=2pt, diag text] (abc) {Models};
        \draw[shorten=4pt, very thin] (def.north west|-abc.north west) -- (def.south east-|abc.south east);}
                                                                                      & \cellcolor{white} G.NMF & \cellcolor{white} KMeans & \cellcolor{white} Lin.Reg & \cellcolor{white} Log.Reg & \cellcolor[HTML]{EFEFEF}Speedups/dataset \\ \hline
\rowcolor{green!5}  
\cellcolor{white} TPC-DI sf=3                                                   & 1.21                             & \cellcolor{red!10} 1      & 1.09                            & 1.23                            & 1.06                                                                                    \\ \hline
\rowcolor{green!5}  
\cellcolor{white} TPC-DI sf=7                                                   & 1.17                             & \cellcolor{red!10} 1      & 1                               & 1.13                            & 1.03                                                                                    \\ \hline
\rowcolor{green!5}  
\cellcolor{white} book                                                   & 1                                & 1                              & 1                               & 1                               & 1                                                                                       \\ \hline
\rowcolor{green!5}  
\cellcolor{white} expedia                                                & 1.24                             & 1.07                           & \cellcolor{red!10} 1       & \cellcolor{red!10} 1       & 1.05                                                                                    \\ \hline
\rowcolor{red!10} 
\cellcolor{white} flight                                                 & 0.52                             & 0.67                           & \cellcolor{green!5} 1       & 0.52                            & 0.71                                                                                    \\ \hline
\rowcolor{red!10} 
\cellcolor{white} lastfm                                                 & 0.94                             & 0.89                           & \cellcolor{green!5} 1       & \cellcolor{green!5} 1       & 0.91                                                                                    \\ \hline
\rowcolor{green!5}  
\cellcolor{white} movie                                                  & 1.69                             & 1.77                           & 2.98                            & 1.78                            & 1.97                                                                                    \\ \hline

\rowcolor{green!5}  
\cellcolor{white} yelp                                                   & 1.06                             & 1.19                           & 1.33                            & 1.28                            & 1.23                                                                                    \\ \hline\hline
\cellcolor[HTML]{EFEFEF}\begin{tabular}[c]{@{}l@{}}Speedups/model\end{tabular} & \cellcolor{green!5} 1.18     & \cellcolor{green!5} 1.09   & \cellcolor{green!5} 1.21    & \cellcolor{green!5} 1.08     & \#                                                                                       \\ 
\end{tabular}
\vspace{-2mm}
\end{table}

\vspace{-3mm}
\section{Related Work}
\label{sec:rw}

\vspace{-2mm}
Early efforts required \emph{manual} design for factorizing specific ML algorithms, with Orion \cite{\orion} addressing linear and logistic regression. 
This was expanded in Santoku \cite{\santoku} to incorporate decision trees, feature ranking, and Naive Bayes. 
While Orion and Santoku focused on inner joins via PK-FK relationship,  F \cite{\f} has extended factorization of linear regression to natural joins. Following F, AC/DC \cite{\acdc} adds support to categorical variables, and LMFAO supports more ML algorithms such as decision trees. JoinBoost \cite{huang2023joinboost} integrated tree models like LightGBM and XGBoost.

 
In a more \emph{automated} manner, 
Morpheus \cite{chen2017towards} proposed a general factorization framework based on PK-FK and join dependencies. HADAD \cite{\hadad} speeds up the rewriting rules of  Morpheus by reordering of multiplication and exploiting pre-computed results. 
Later, Trinity \cite{\trinity} extends Morpheus with compatibility to 
multiple programming languages and LA tools. 
MorpheusFI \cite{\morpheusfi} added support to non-linear feature interactions. Non-linear models such as 
Gaussian Mixture Models and Neural Networks are studied in \cite{DBLP:conf/icde/ChengKZ021}.

Ilargi's main contribution is not on extending ML algorithms. We expand the usability of factorized learning by a GPU-compatible representation. We take a more critical inspection of the speedup of factorization compared to materialization by utilizing DI metadata. We also discovered a third factor, hardware configuration, which has an impact on the decision boundary. We are currently expanding Ilargi to more ML models such as tree models.

\vspace{-4mm}
\section{Conclusion}
\label{sec:con}
\vspace{-2mm}
In this paper, we have proposed \emph{Ilargi}, an approach that leverages matrix-represented DI metadata to enable GPU-compatible factorization of model training across disparate data sources. With the unified LA representation, Ilargi efficiently trains models using factorization on both CPUs and GPUs. Moreover, we introduced a tree-boosting estimator to navigate the complex decision-making process between materialization and factorization, combining data characteristics, ML model training complexity, and hardware configuration.
Our experimental results on both synthetic and real-world data demonstrate that our GPU-compatible factorized model training can achieve speedups of up to 8.9x, attributing to the high parallelism of GPUs. 
With our estimator, the batch model training workload can be accelerated by up to 24\% on both CPUs and GPUs consistently, demonstrating the benefits of incorporating hardware specifications into our estimator.

\vspace{-3mm}
%
%
%

%
%
%
\bibliographystyle{splncs04}
\bibliography{reference}
%




\end{document}